\documentclass[]{spie}  %>>> use for US letter paper
\newcommand{\argmin}{\mathop{\rm arg~min}\limits}
\newcommand{\hzy}[1]{\textcolor{black}{#1}}
\usepackage{amsmath,amsfonts,amssymb}
\usepackage{graphicx}
\usepackage{url}
\usepackage{cite}
\usepackage{amsmath,amssymb,amsfonts}
\usepackage{algorithmic}
\usepackage{graphicx}
\usepackage{textcomp}
\usepackage{xcolor}
\usepackage{stfloats} 
\usepackage{bm}
\usepackage{subfigure}
\usepackage{booktabs}
\usepackage{diagbox} 
\usepackage{multirow}
\usepackage{makecell}
\usepackage{longtable}
\usepackage{cite}
\usepackage[colorlinks=true, allcolors=blue]{hyperref}

\title{Interactive 3D Character Modeling from 2D Orthogonal Drawings with Annotations}

\author[a]{Zhengyu Huang}
\author[a]{Haoran Xie}
\author[b]{Tsukasa Fukusato}
\affil[a]{Japan Advanced Institute of Science and Technology, Ishikawa, Japan}
\affil[b]{The University of Tokyo, Tokyo, Japan}

\authorinfo{Further author information: (Send correspondence to Tsukasa Fukusato)
\\Zhengyu Huang: E-mail: huang.zhengyu@jaist.ac.jp
\\Haoran Xie: E-mail: xie@jaist.ac.jp
\\Tsukasa Fukusato: E-mail: tsukasafukusato@is.s.u-tokyo.ac.jp, Telephone: 81-3-5841-4109}
\begin{document} 
\maketitle

\begin{abstract}
%\footnote{xx}
%We propose an interactive 3D character modeling approach from orthographic drawings (e.g., front and side views) based on 2D-space annotations. 
%First, the system builds partial correspondences between the input drawings and generates a base mesh with sweeping splines according to edge information in 2D images. 
%Next, users annotates the desired parts on the input drawings (e.g., eye and mouth) by using two-type strokes, named addition and erosion, and the system re-optimizes the shape of the base mesh. %by using the annotations. 
%By repeating the 2D-space operations (i.e., revising and modifying the annotations), users can design a desired character model. To validate the efficiency and quality of our system, % a user study is conducted comparing our approach with a conventional workflow such as 3D-space operations. 
%we also verify the generated results with state-of-the-art methods to show the quality of our system. 
%we compared the generated results with those from state-of-the-art methods.
%we verify the generated results with state-of-the-art methods.

We propose an interactive 3D character modeling approach from orthographic drawings (e.g., front and side views) based on 2D-space annotations. 
First, the system builds partial correspondences between the input drawings and generates a base mesh with sweeping splines according to edge information in 2D images. 
Next, users annotates the desired parts on the input drawings (e.g., the eyes and mouth) by using two type of strokes, called addition and erosion, and the system re-optimizes the shape of the base mesh.
By repeating the 2D-space operations (i.e., revising and modifying the annotations), users can design a desired character model. To validate the efficiency and quality of our system,
we verified the generated results with state-of-the-art methods.
\end{abstract}

% Include a list of keywords after the abstract 

\section{INTRODUCTION}
\label{sec:intro}  % \label{} allows reference to this section
In the animation and game industries, when modeling new 3D characters (or objects), artists first draw orthographic
views of them. 
%When modeling a new character, it is commonly to sketch 2D images in front view, back view and side view on the initial stage for artists. 
However, it is cumbersome and time consuming converting 2D drawings into 3D models manually because 3D modeling with specialized tools (e.g., Maya and 3DMAX) require professional knowledge and those user interfaces are not as intuitive as 2D drawing.

Although several sketch-based modeling methods have been proposed for 3D content creation~\cite{BhattacharjeeC20}~\hspace{-6pt}, it is still a challenging issue to represent characteristics of character drawings --- there is a gap between the generated results and professional 3D modelings. 
To solve this issue, we proposed a user interface to easily and efficiently design such characteristics on 3D shapes with the help of 2D annotations. Leveraging orthogonal views, our system can faithfully reconstruct 3D models from drawings. 
The main contribution of this paper is to provide a novel user-friendly workflow for designing 3D models from 2D drawings with sketch-like annotations, which eliminates the need for complex 3D operations. 
\begin{figure}[t]
\centering
\includegraphics[width=0.9\linewidth]{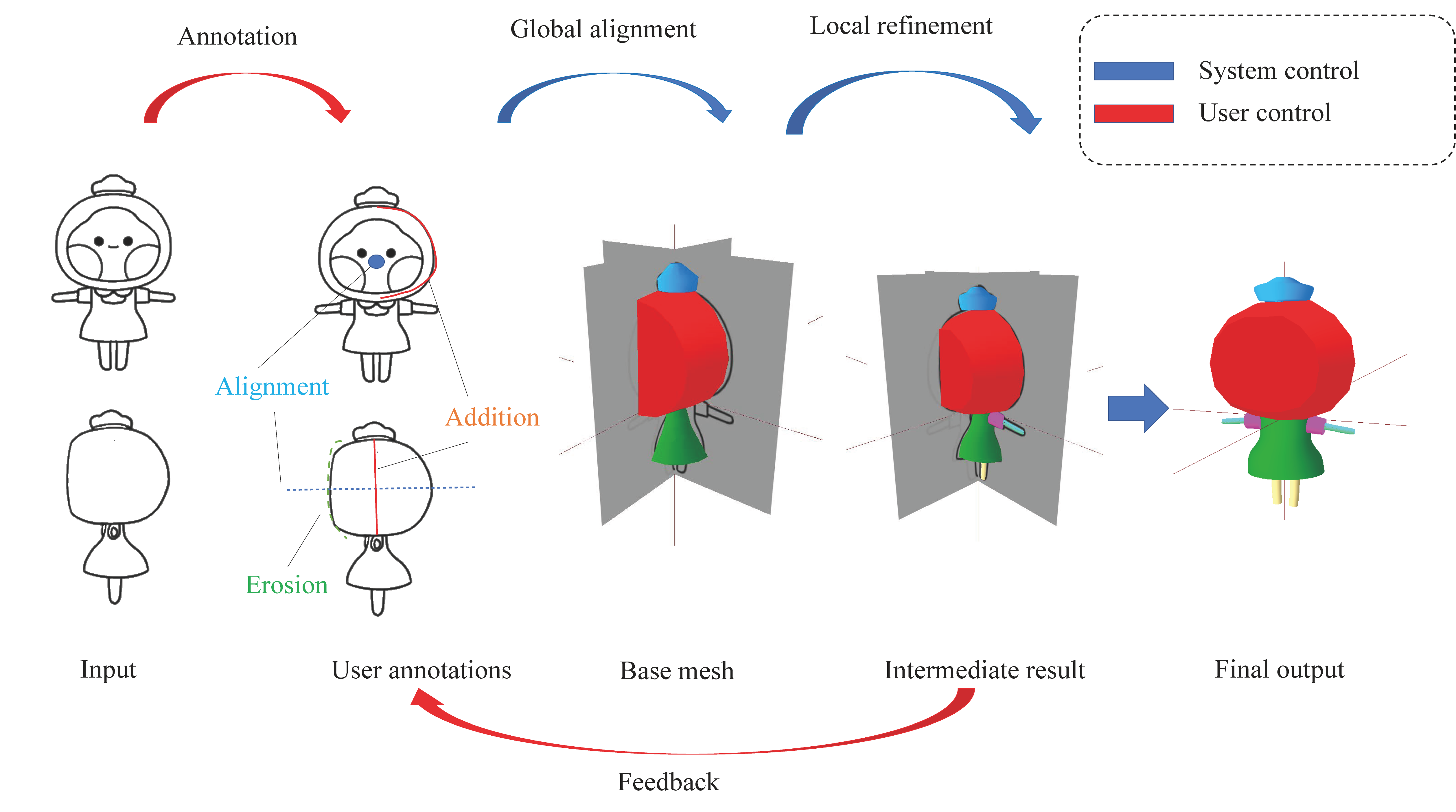}    
\caption{Overview of the proposed system.}
\label{fig:1}
\end{figure}
\section{Related Work}
Sketching is a form of artistic expression that is highly abstracted from the real world that has been used in various graphics applications, such as normal map editing~\cite{he2021}, flow design~\cite{sketch2vf19}, and portrait drawing~\cite{dualface2021}. As input for 3D modeling, 2D sketch has ambiguity problem in free-form drawing. 
To solve this issue, Teddy~\cite{TeddyIgarashiMT99} was proposed as one of the earliest free-form sketch modeling user interface. Once 3D viewpoints are determined by users, smooth surfaces are generated by interpolating the curves extracted from users' sketches. Several interpolation functions for sketch modeling~\cite{KarpenkoH06,Repouss_JoshiC08,Matisse_BernhardtPCB08} were proposed to improve the results. However, these approaches usual tended to get over-smoothed surfaces.
Some approaches, such as BendSketch~\cite{BendSketch:LiPLTSW17} offered a solution for this issue.

Another popular approach to sketch-based modeling is data-based learning. By analyzing a massive number of sketch-model pairs, methods of this type can generate an accurate 3D model from a user's simple sketch. %With development of 3D deep learning.
Smirnov et al.~\cite{Smirnov2019} applied Coons patches to learn shape surfaces, but their method is limited to generating smooth shapes. 
Sketch2CAD allows users to create objects incrementally with sketches, which were inferred to CAD instructions by convolutional neural networks~\cite{Li:2020:Sketch2CAD}.
%Du et al. \cite{DuZNHCYL21} decomposed 3D objects into parts and adopt implicit learning with low-resolution mesh extraction for distinct parts to guarantee accuracy.
SimpModeling provides a sketching system for animalmorphic-head modeling which can generate details of a head from sketches by pixel-aligned implicit learning.

SketchModeling~\cite{LunGKMW17} is very relevant to our work, which is also attempting to reconstruct character models from multi-views of sketch images, though it is an automatic approach. With an encoder-decoder U-Net architecture, SketchModeling can get depth maps and corresponding normal maps from input sketches, optimize the point cloud by merging these views, and obtain complete 3D models. \hzy{Inspired by structured annotations~\cite{GingoldIZ09}, which is a generalized-cylinder-based modeling for a single view, we propose a user interface to easily and efficiently design characteristics on 3D shapes with fewer types of annotations leveraging the orthogonal views.}

\section{User interface}
\begin{figure}[t]
\centering
\includegraphics[width=0.9\linewidth]{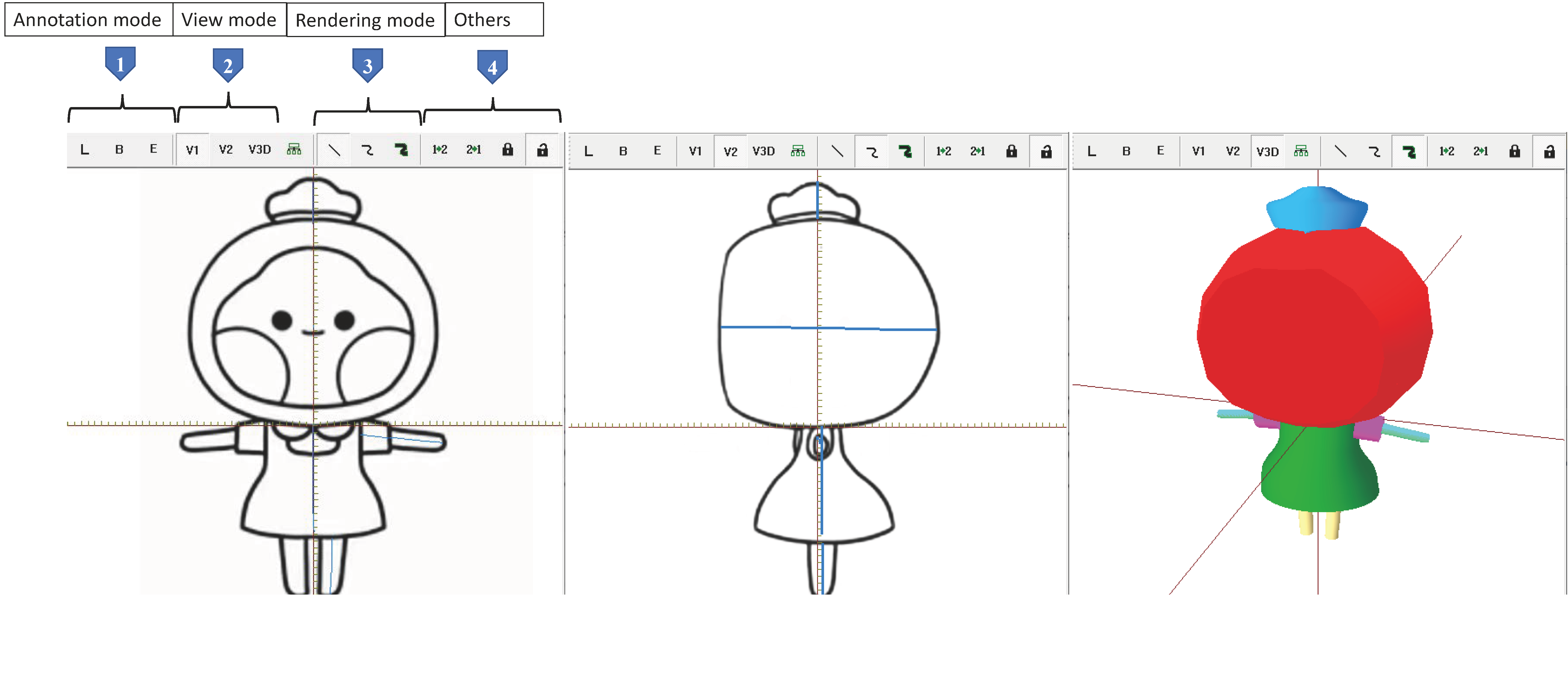} 
\caption{
User interface in the proposed system. A user models each part of the character by annotating on both views and viewing the results in 3D view. 
}
\label{fig:ui}
\end{figure}

In this section, we describe how users interact with the proposed two-stage user interface (see Fig.~\ref{fig:ui}) to model a character with annotations. The tool bar consists of four parts: (1) Annotation mode, including local mode (alignment annotation addition and edge/background marking for corresponding alignment annotation), addition annotation boundary addition(B), and erosion annotation addition (E) from left to right; (2) View mode, including 2D front view (V1), 2D side view (V2), 3D view (V3D) and selected-annotation-only mode; (3) Rendering mode for annotations, including drawing as segments, drawing as curve and drawing as generated cylinder; (4) Other options. including relocation a selected annotation from V1 to V2 (and V2 to V1), a lock button and a unlock button shows whether or not adopting epipolar constraint from the other view as a reference when relocating.

\subsection{Annotation Tool}
Since input 2D orthogonal drawings often do not provide complete information for 3D modeling, our system allows the user to freely draw annotations that are not limited by edge information.  %provided
The user can draw brief strokes in either front view or side view by inserting key points on the canvas with a mouse-click operation, and each stroke can be labelled as alignment, addition, or erosion. Then, the system automatically generates corresponding strokes in the other view with the same label. The user is allowed to edit strokes in editing mode to calculate correct 3D coordinates of strokes in the 3D view. In contrast, with the eraser tool, the user clicks on a stroke, and the system deletes the selected stroke. Moreover, the undo shortcut (``Z'' key) can delete the last stroke from the stroke list.
Note that our system can also load or export the user-drawn annotations by pushing down the ``Load''(``L'') or ``Save''(``S'') key on the keyboard. 

\subsection{Editing Tool}
In order to facilitate repeated modification and find the appropriate correspondence between annotations in two views, the system provides an editing function. In 2D-view mode (V1 or V2), any vertex position of the corresponding annotation can be modified by selecting any visible curve. In editing mode, the system allows the user to generate polar constraints using the corresponding annotations of another view, reducing the 2D editing of the vertices to 1D editing.

\section{Overview}
Figure~\ref{fig:1} shows an overview of our sketch-based character modeling system.
After a character design is input, users can draw corresponding annotations one by one in both front and side views under the epipolar constraint. Users can also add extra edge information that is invisible in the original images by sketching for each part of the model with the help of our region-based boundary extraction.
Once the user's annotations are completed, the corresponding lines from alignment annotations (blue strokes) are extracted as hard constraints, and those lines marked with the other two annotations are excluded to revise relationships between the two views of the sketches and calculate more precise coordinates of 3D points for base mesh. After the base mesh is determined, cloud points will be sampled from edge information according to the addition annotations (orange strokes) and erosion annotations (green strokes) as constrains and optimization-based surface fitting is conducted to generate a smooth surface. 

%\subsection{Data Structure}

\subsection{Alignment-Based Global Modeling (Base-Mesh Generation)}
%Todo fig for UI, base-mesh generation, refinement ,examples
\begin{figure}[t]
\centering
\includegraphics[width=0.8\linewidth]{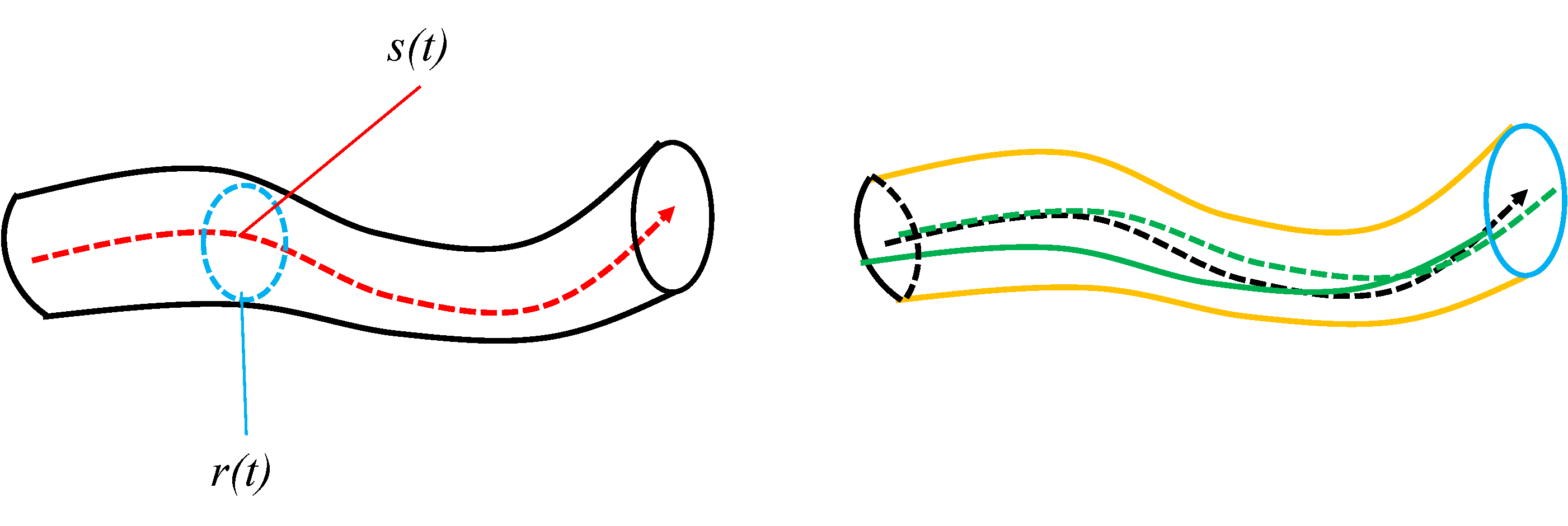} 
\caption{
Generalized cylinders. Candidate boundaries which would be projected into 2D views are colorized in right column. Three type boundaries are orthogonal from one to each others.
}
\label{fig:gc}
\end{figure}

In the first step, generalized cylinders are generated as a base mesh according to corresponding alignment annotations in two views. Each part of the character $P$ is modelled separately. Here, any annotation $A$ is restored as a series of ordered $n$ key points $\{\bm{p_0,p_1,...,p_{n-1}}\}$, belonging to a single part and represented as a Hermitian curve. % of the proposed system
In this system, alignment annotations are mainly used to represent the skeleton or center of gravity of modeling parts and to correct the 3D position of a specific curve in some cases.

\noindent
\textbf{Primitives}. A typical generalized cylinder is shown in Fig.~\ref{fig:gc}, consisting of a skeleton curve $\bm s(t) (t \in [0,1])$, and a cross-section radial distance function $\bm r(t)$. Here, different types of candidate boundaries which may be projected onto two views are marked with different colors.

\noindent
\textbf{Camera model and epipolar constraint}.
In $\mathbb{R}^3$ space, extrinsic parameters of a camera for 3D reconstruction can be denoted
by a translation vector $T$ and a $3 \times 3$ rotation matrix $\bm R=[\bm{R}_{0},\bm{R}_{1},\bm{R}_{2}]^T$. Here $f$ denotes the focal length of the camera. 
Given a set of 3D points $P$, the set of points projected onto the front view is $Q_1$, and the set of points projected onto the side-view is $Q_2$. If $\bm p \in P $, $\bm q_1 \in Q_1 $ and $\bm q_2 \in Q_2 $ correspond to each other and the camera parameters of two views are ($\bm R_1,\bm T_1,f_1$) and ($\bm R_2,\bm T_2,f_2$), respectively, then, $\bm q_1$ and $\bm q_2$ can be expressed with following equations:
%the correspondence relation between 
\begin{equation}
\left\{\begin{matrix}
\bm q_{1x} &= f_1 \frac{ \bm{R}_{10} \cdot \bm v+T_{1y}}{ \bm{R}_{12} \cdot \bm{v} + T_{1z}}\\
\bm q_{1y} &= f_1 \frac{ \bm{R}_{11} \cdot \bm v+T_{1y}}{ \bm{R}_{12} \cdot \bm{v} + T_{1z}}\\
\bm q_{2x} &= f_2 \frac{ \bm{R}_{20} \cdot \bm v+T_{2y}}{ \bm{R}_{22} \cdot \bm{v} + T_{2z}}\\
\bm q_{2y} &= f_2 \frac{ \bm{R}_{21} \cdot \bm v+T_{2y}}{ \bm{R}_{22} \cdot \bm{v} + T_{2z}}\\
\end{matrix}\right.
\label{eq1}
\end{equation}
%where 
%As the two 2D views are orthogonal to the y-axis and $f_1=f_2$ in our case, epipolar constraint derived from Equation. 1 can be simplified as:
%If a 2D point obtained by projecting 3D point $v \in V $ one the image of one view is u, the 2D point obtained by projecting in the other view of the skeleton point is u'. As the two camera parameters and the skeleton point u in the photograph of the first viewpoint are know, the point u and u' must be on its polar line, and the polar line satisfies the following equation:

\noindent
Since the two views are orthogonal to the y-axis, %and the 2D alignment annotations 
 $\bm q_1$ and $\bm q_2$ are located exactly at orthogonal planes. We also have $f_1=f_2$, $T_{1z}=T_{2z} = 0.0$, and $T_{1y}=T_{2y}$. Then, the epipolar constraint derived from Equation.~\ref{eq1} can be simplified as:
\begin {equation} 
\bm q_{1y}=\bm q_{2y}%+(T_1.y-T_2.y)
\end {equation}
Thus, the corresponding 3D positions can be calculated correctly.
For instance, support a point $\bm p$ with coordinates $(x_1,y_1)$ in the front view and $(x_2,y_2)$ in the side view. Since $y_1=y_2$ epipolar line constraint, its 3D coordinates are $(x_1,y_1,-x_2)$.
 Although we are using only this special case, the polar constraints can be naturally extended to reduce the workload of a multi-view alignment.
 
\noindent
\textbf{Base mesh generation.} Our system uses the edge information of the image and the relative position of alignment annotations to automatically calculate the preliminary boundary of 2D generalized cylinders of for each view. For any point on the skeleton curve $\bm s(t)$ of an alignment annotation where $t=t'$, the nearest intersection points in two directions between its vertical line ends and edges can be found as an initial boundary $\bm b(t)$ of cross-section $\bm r(t)$ . If edges $E$ denotes a set of world coordinates converted from edge pixels in input images and function $D(E,\bm b(t))$ denotes the distance between $E$ and $\bm b(t)$, the formula can be described as follows:
\begin{equation}
\left\{\begin{matrix}
\bm b(t) = \argmin(D(E,\bm b(t)))\\
D(E,\bm b(t)) = min_{\bm e \in E} \{ \left \|\bm b(t)+\bm s(t)-\bm e \right \|_2+ \left \| \bm b(t)\right \|_2 \}
\end{matrix}\right.
\label{eq3}
\end{equation}
Note that the mesh generated by this method tends to fail for edges that are close to parallel. The main reason for this is the need to discretize the mesh when generating it, so parameter $t$ for $s(t)$ has a certain interval. This weakness will be overcome in the following local refinement step.

\subsection{Local Refinement with Annotation Constrains in Two Orthogonal Views}
The next step is refining the base mesh with optimization. Here, we introduce addition annotations and erosion annotations, as shown in Fig.~\ref{fig:1}, to realize this objective: addition annotations $A_{a,t}$ define boundaries for cross-sections along a skeleton curve, while erosion annotations $A_{e,t}$ modify shapes of end-caps of generalized cylinders. 
Note both annotations only work when they attached to an alignment annotation specified by the user.

These annotations should be converted to $\bm b(t)_k$, denoting a boundary function with a type of $k$ ($k \in K$).
$K$ is a set of types of boundary, and in our case K={0, 1, 2} denoting a blue curve (cross-section contour), orange curve, and green curve in the right column of Fig.~\ref{fig:gc}, respectively.
All reconstructed parts of character should have minimized errors between the visible contours and annotations constraints when project back to 2D views. %Support $\bm b(t)_k$ denotes a boundary function with type of k, 
The objective function $F(B, E, A)$ in this step can be summarized as:
 
 \begin{equation}
\left\{
\begin{matrix}
B=\argmin F(E, A)\\
min F(E, A)=\sum_{\bm b(t)\in A, k\in K}{D(E,\bm b(t)_k)} %\sum\limits
\end{matrix}\right.
\label{eq5}
\end{equation}
where $B$ is a set of boundaries extracted from annotations; edges $E$ and function $D(E,\bm b(t))$ are the same as the one in Equation.~\ref{eq3} described. Once $B$ is determined, the base mesh in the global step can be refined as follows:

\noindent
\textbf{Cross-section modeling vertical to the skeleton direction of generalized cylinder}. 
When $\bm b(t)_0 \notin B$, there are only at most four constraint points for each cross-section. In this case, the system would fit an ellipse by regarding these constraint points as its poles. Specially, only one constraint point in the cross-section means this cross-section is a circle.

If $\exists t=t'$ let $\bm b(t)_0 \in B$, cross-sections between these addition annotation boundaries will be calculated with cubic B-spline interpolation.%Hermite interpolation. 
As the cross-sections along to the skeleton direction of generalized cylinder have been calculated, 
the side surface of the generalized cylinder matching with input and user-defined boundaries can be generated.
%\textbf{Modeling along to the skeleton direction of generalized cylinder.} 

\noindent
\textbf{End-caps of generalized cylinder. }
If there is an erosion annotation for a generalized cylinder, the end-caps surfaces will be deformed with Laplacian-based editing~\cite{LaplacianEditingAuTLF06} according to the annotation's shape. Otherwise, surfaces at the ends of the generalized cylinder are planes.

The main idea of this step is to use two kinds of annotations to obtain constraints for a generalized cylinder. Therefore, this step can not only perform boundary refinement, but also improve the effectiveness of boundary classification, which in turn improves modeling efficiency and accuracy.

\section{Result and Discussion}

\begin{figure}[t]

\begin{minipage}{0.3\linewidth}\centering
\includegraphics[width=\textwidth]{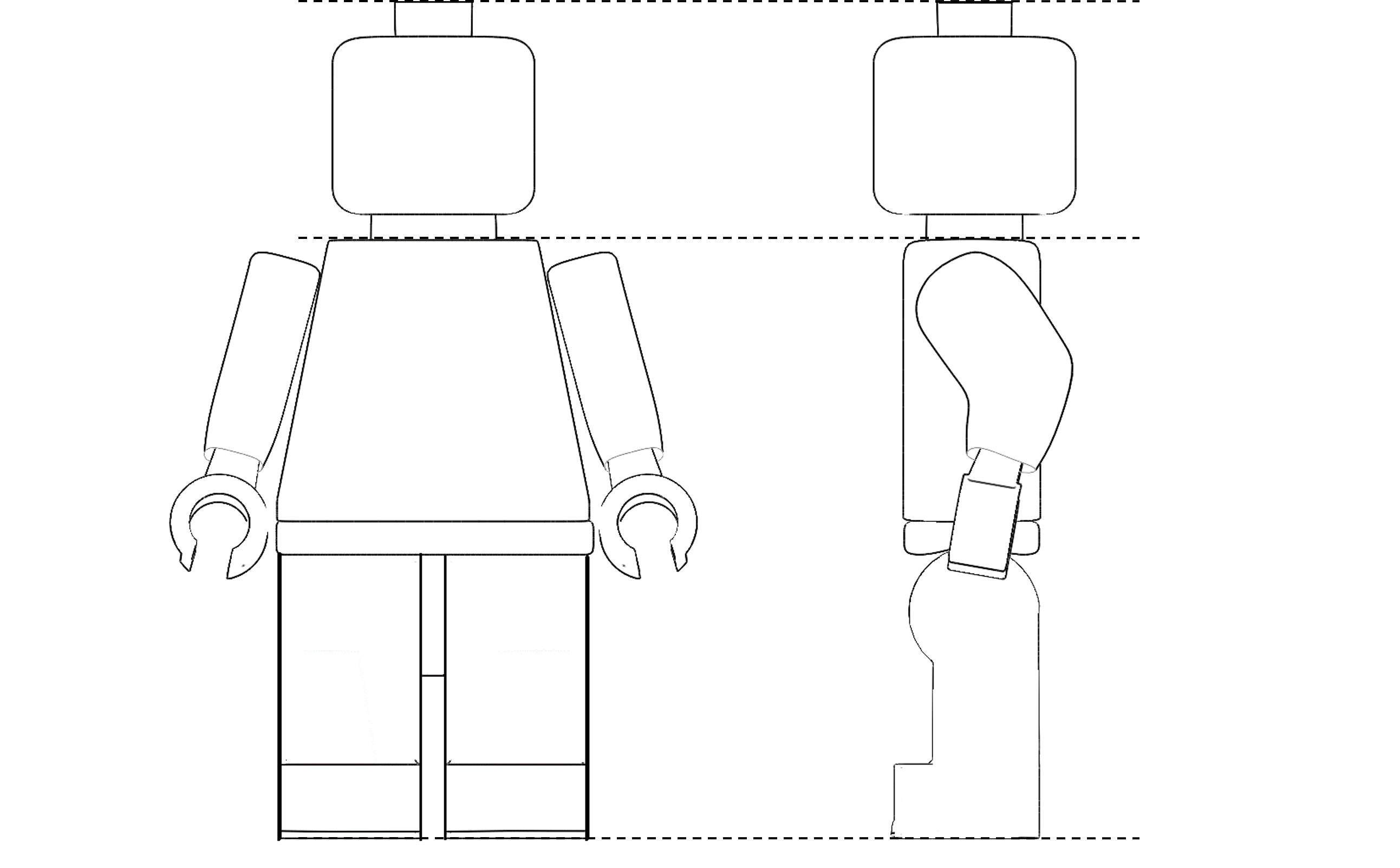}\\
\includegraphics[width=\textwidth]{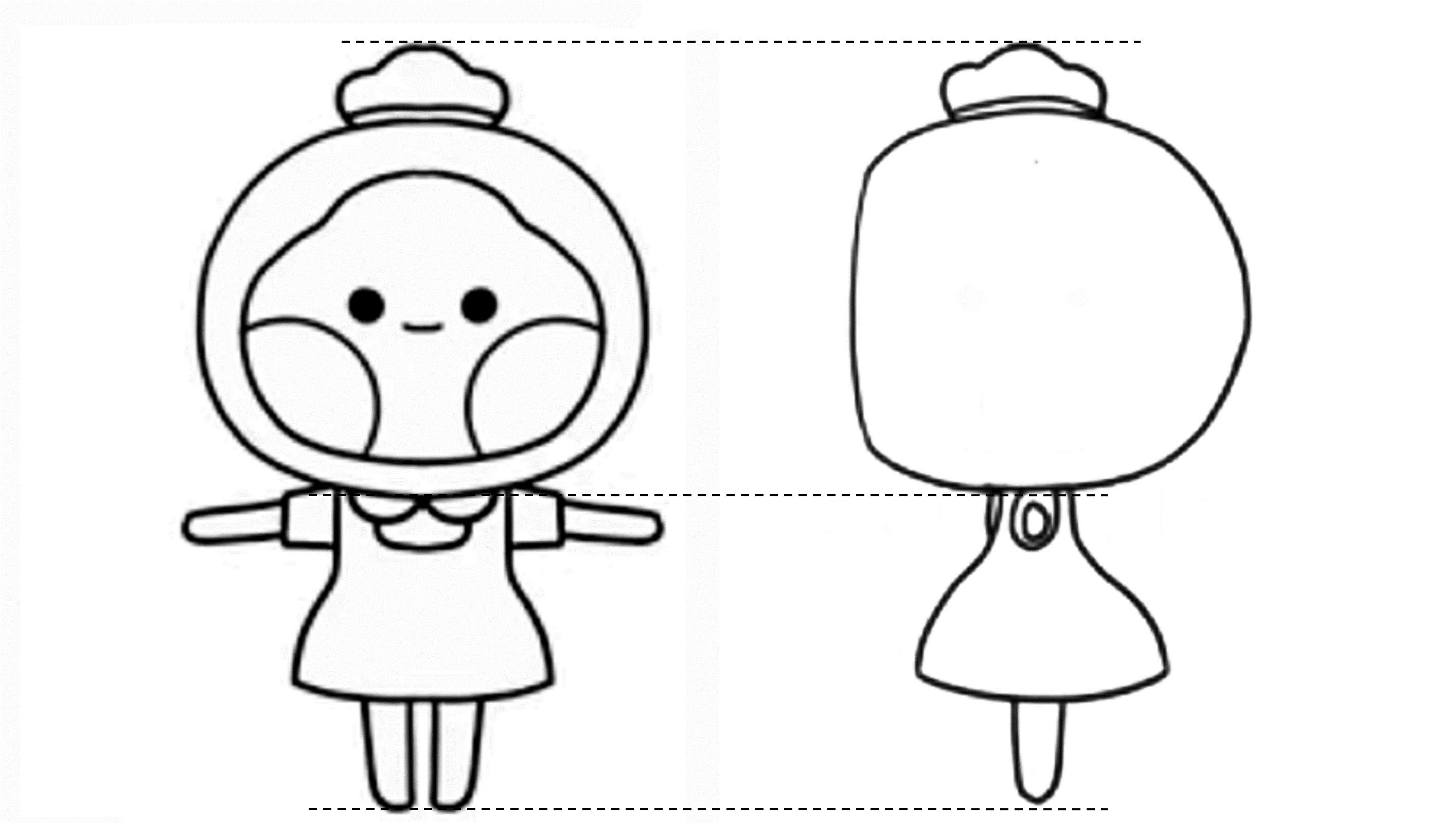}\\
\vspace{13.5pt}
\subfigure[Input images]{\quad\quad\quad\quad\quad\quad\quad\quad}
\vspace{-13.5pt}
\end{minipage}
\begin{minipage}{0.3\linewidth}\centering
\includegraphics[width=0.4\textwidth]{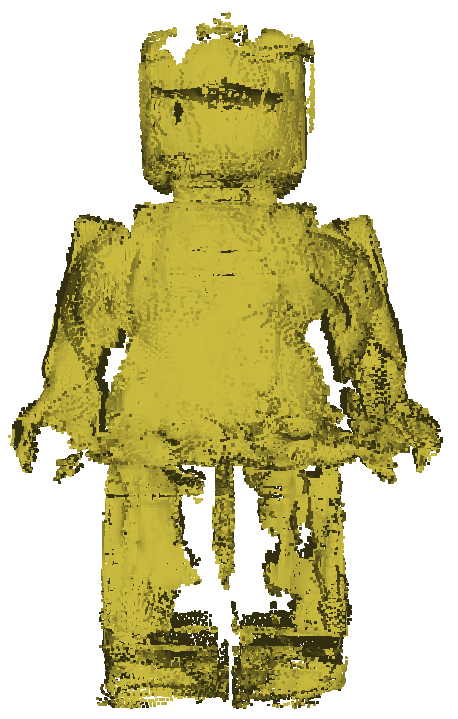}\includegraphics[width=0.48\textwidth]{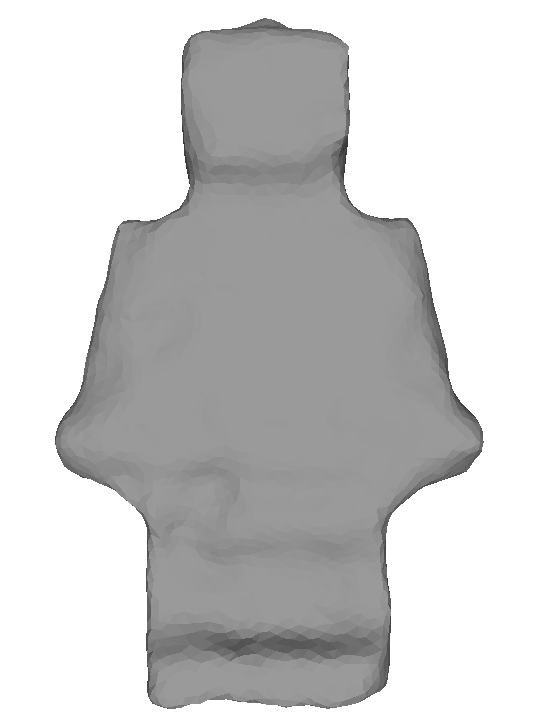}\quad 251mins\\
\includegraphics[width=0.38\textwidth]{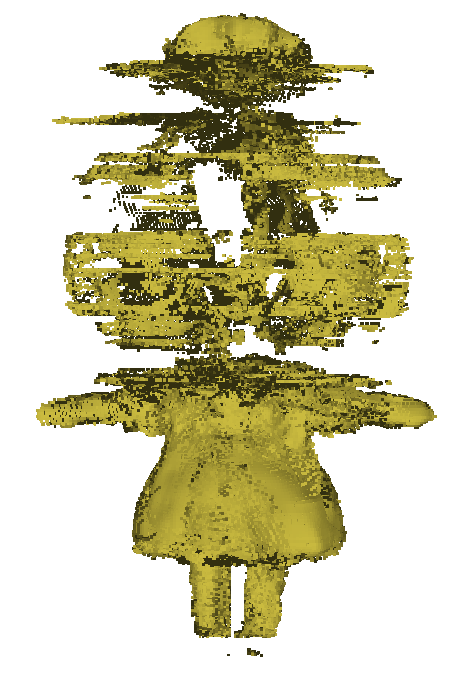}\includegraphics[width=0.42\textwidth]{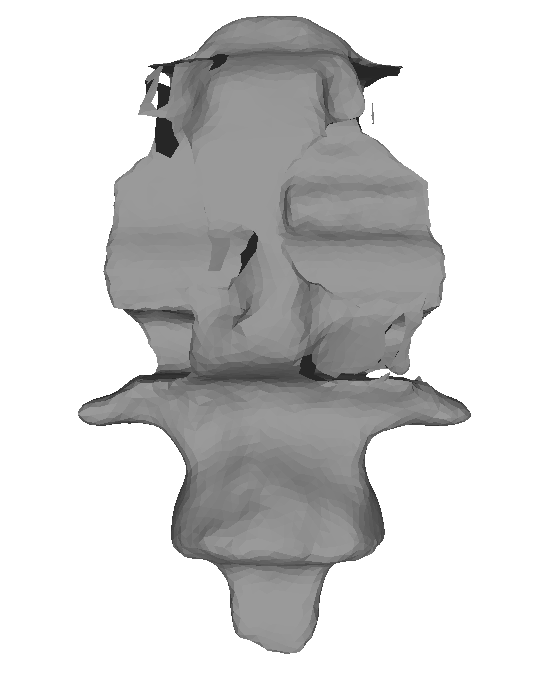}\quad 221mins\\
\subfigure[SketchModeling]{\quad\quad\quad\quad\quad\quad\quad\quad}
\end{minipage}
\begin{minipage}{0.3\linewidth}\centering
\includegraphics[width=0.436\textwidth]{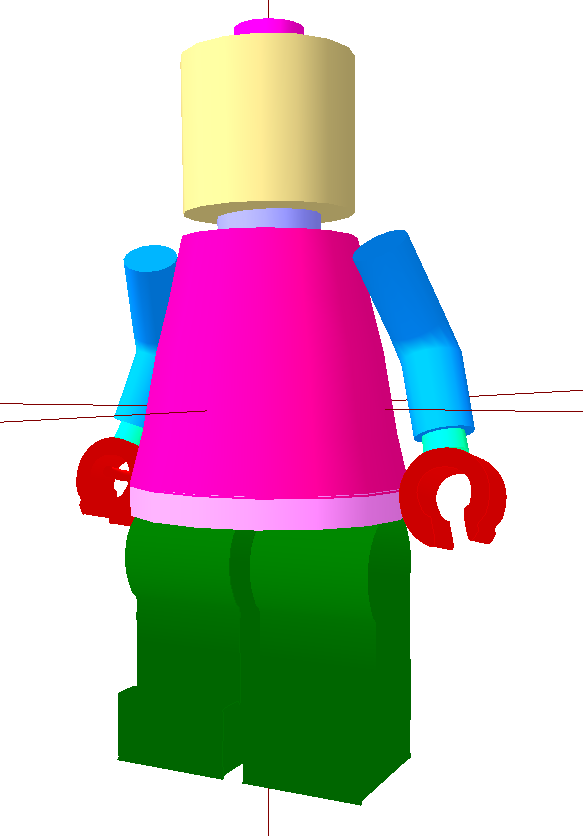}\\\quad 8-parts, 6mins\\
\includegraphics[width=0.51\textwidth]{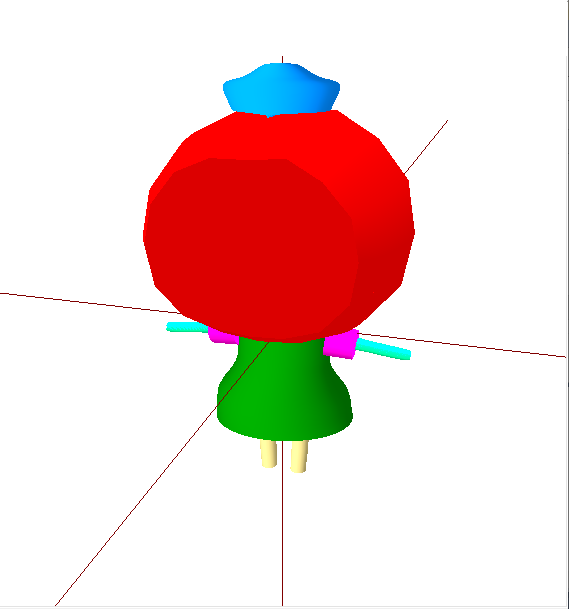}\\\quad 6-parts, 8mins\\
\subfigure[Ours]{\quad\quad\quad\quad\quad\quad\quad\quad}
\end{minipage}
\caption{The models created in the comparison study with SketchModeling\cite{LunGKMW17} and our proposed system.}
\label{fig:res}
\end{figure}

In our implementation, the system was programmed in C$++$ as a real-time drawing application on the Windows 10 platform. 
A workstation with Intel Core i5-8400, 2.80GHz 2.80GHz, NVIDIA RTX2070 GPU, and 16GB RAM was used as the testing computing environment. Figure~\ref{fig:res} shows 3D modelling results with our system comparing with SketchModeling~\cite{LunGKMW17}. Both the final cloud points and mesh of the first-row character do not match well with the input sketch \hzy{after hours of calculations}, which means this learning-based method failed to predict the position and depth-map of some parts of the characters, while models generated with our system were more faithful to the original design drawings \hzy{in less time by modelling 
several independent parts}.
%The user editting time is XX mins and reconstruction calculation time is XX seconds.

Unlike the template-based method, the proposed system enables users to make character models without careful parameter tuning.
%intuition and parameter adjustment is not required. 
By comparing our results with results of state-of-the-art methods, %and collecting feedback from user study, 
we verified that the proposed system could improve the quality of 3D character models with simpler but more intuitive operations. 
The current system focuses mainly on the 3D shapes reconstruction process, so texture mapping and complex structure modeling might be a good topic for future research. 
%The current system focuses mainly on the 3D shapes reconstruction process, so texture mapping and complex structure modeling should be a future endeavor.
%As a limitation, 

\acknowledgments % equivalent to \section*{ACKNOWLEDGMENTS}   
%We thank the anonymous reviewers and the editor for their insightful comments that improved this manuscript. Other acknowledgements removed for review.
This research was supported by the Kayamori Foundation of Informational Science  Advancement, JSPS KAKENHI JP20K19845, and JP19K20316.
%
%This unnumbered section is used to identify those who have aided the authors in understanding or accomplishing the work presented and to acknowledge sources of funding.  

% References
\bibliography{report} % bibliography data in report.bib
\bibliographystyle{spiebib} % makes bibtex use spiebib.bst

\end{document}